\RequirePackage{fix-cm}
\documentclass[review]{elsarticle}

\usepackage{hyperref}

\usepackage{pgf-pie}
\usepackage{subfiles}
\usepackage{graphicx}
\usepackage[utf8]{inputenc}
\usepackage{comment}
\usepackage[english]{babel}
\usepackage{textcomp}
\usepackage{adjustbox}
\usepackage{indentfirst}
\usepackage{mathtools}
\usepackage[usestackEOL]{stackengine}
\usepackage{ltxtable}

\usepackage{amsmath}
\usepackage{amsfonts}

\usepackage{algorithm}
\usepackage{algorithmicx}
\usepackage{algpseudocode}

\usepackage{textcomp}
\usepackage{mathpazo}
\usepackage{tgpagella}
\usepackage{color,soul}
\usepackage{dirtytalk}
\usepackage{tcolorbox}
\usepackage{bm}
\usepackage{graphicx,subcaption,chngpage,calc}
 \usepackage[math]{cellspace}
  \usepackage{ltablex}
\usepackage{xtab, lipsum, array}
\usepackage{caption}
\usepackage{multirow}
\usepackage{tabularx,ragged2e,booktabs,caption}
\newcolumntype{L}{>{\raggedright\arraybackslash}X}

\stackMath

\DeclareMathSizes{12}{17.28}{9}{7}
\usepackage{color}
\usepackage{multirow, makecell}
  \usepackage{ltablex}

\usepackage{url}

\usepackage{tabularx}
\usepackage{pgfplots}
\pgfplotsset{compat=1.14}
\usepackage{pgfplotstable}
\usepackage{sfmath}
\usepackage{array, booktabs, makecell}
\usepackage{color}
\usepackage{csquotes}
\usepackage{url}
\usepackage{comment}
\usepackage{tikz}
\usepackage{balance}
\usepackage{listings}
\usepackage{breqn}
\usepackage{booktabs} % used for \toprule in tables
\usepackage{soul} % highlighting
\usetikzlibrary{fit} %% Used for putting dotted box in image
\usetikzlibrary{positioning}
\usetikzlibrary{arrows}
\usetikzlibrary{shapes.multipart}
\usepackage{multirow} 
\usepackage{multicol}
\usepackage{varwidth} % Double column tables

\usepackage{tikz,xcolor,hyperref}

\definecolor{Gray}{gray}{0.80} % the lower the #, the darker it gets

\definecolor{bblue}{HTML}{4F81BD}
\definecolor{rred}{HTML}{C0504D}
\definecolor{ggreen}{HTML}{9BBB59}
\definecolor{ppurple}{HTML}{9F4C7C}

\pgfplotsset{
/pgfplots/my legend/.style={
legend image code/.code={
\draw[thick,black](-0.05cm,0cm) -- (0.3cm,0cm);%
   }
  }
}

\usepackage{mathtools}
\usepackage{amsmath}

\makeatletter
\def\BState{\State\hskip-\ALG@thistlm}
\makeatother

\begin{document}
\begin{frontmatter}

\title{Distil-xLSTM: Learning Attention Mechanisms through Recurrent Structures}

\author[First]{Abdoul Majid O. Thiombiano}
\ead{abdoulmajid.ousseinithiombiano@fsm.rnu.tn}

\author[First,Second]{Brahim Hnich}
\ead{brahim.hnich@fsm.rnu.tn}

\author[Third]{Ali Ben Mrad}
\ead{a.benmrad@qu.edu.sa}

\author[Fourth]{Mohamed Wiem Mkaouer}
\ead{mmkaouer@umich.edu}

\address[First]{FSM, University of Monastir, Monastir, 5000 Tunisia}
\address[Second]{CES Lab, ENIS, University of Sfax, Sfax, 3038 Tunisia}
\address[Third]{Department of Computer Science, College of Computer, Qassim University, Buraydah, Saudi Arabia}
\address[Fourth]{University of Michigan-Flint, MI, USA}

\date{Received: DD Month YEAR / Accepted: DD Month YEAR}

\begin{abstract}
    The current era of Natural Language Processing (NLP) is dominated by Transformer models. However, novel architectures relying on recurrent mechanisms, such as xLSTM and Mamba, have been proposed as alternatives to attention-based models. Although computation is done differently than with the attention mechanism\footnote{In this paper, attention refers to self-attention (\cite{vaswani_attention_2023})} mechanism, these recurrent models yield good results and sometimes even outperform state-of-the-art attention-based models. In this work, we propose Distil-xLSTM, an xLSTM-based Small Language Model (SLM) trained by distilling knowledge from a Large Language Model (LLM) that shows promising results while being compute and scale efficient. Our Distil-xLSTM focuses on approximating a transformer-based model attention parametrization using its recurrent sequence mixing components and shows good results with minimal training.
\end{abstract}


%\keywords{Bug report, bug assignment, learning to rank, software quality}

%\begin{IEEEkeywords}
%Bug report, Bug assignment, learning to rank, software quality,
%\end{IEEEkeywords}

\end{frontmatter}

\section{Introduction}

Training large-scale LLMs is now a mainstream task in the research community (\cite{jiang_mistral_2023, touvron_llama_2023}). However, a spark of interest appeared in the training of these models on a smaller scale and led to the birth of models like Phi (\cite{abdin_phi-3_2024}). Despite their large adoption and their proven efficiency in different tasks such as coding (\cite{hui_qwen25-coder_2024}), or even tasks that required the combination of vision capabilities and language understanding (\cite{wang_qwen2-vl_2024}), transformer-based models presented a fundamental computation issue due to the quadratic complexity of the attention mechanism (\cite{vaswani_attention_2023}). 

Recurrent architectures like Mamba (\cite{mamba2}), a state-space model (SSM), and xLSTM (\cite{beck_xlstm_2024}) have been proposed as alternatives to the Transformer architecture. With linear scaling, these models have shown promising results in various tasks (\cite{alkin_vision-lstm_2024, anthony_blackmamba_2024, ren_samba_2024}). 

In an effort to reduce the computational complexity of language models, \cite{katharopoulos_transformers_2020} demonstrated that transformer-based models with causal attention can be reformulated as recurrent neural networks (RNNs). This finding opens the door to exploring lightweight and efficient alternatives to transformer architectures while retaining their expressive power. However, the challenge remains: can we capture the intricate dynamics of attention mechanisms within a recurrent framework without sacrificing performance?

Motivated by the growing demand for scalable models that operate efficiently in resource-constrained environments, we aim to approximate the attention mechanism of transformer models using a recurrent architecture. To this end, we leverage the xLSTM, a novel recurrent architecture featuring enhanced memory mixing and parallel computation, to emulate attention-like behavior effectively.

To achieve this, we adopt a computational and parametric approach based on knowledge distillation (\cite{hinton_distilling_2015}), enabling the transfer of representational capabilities from an attention-based model to a recurrent one. 
%\sout{This led us to experiment with a different kind of knowledge distillation (\cite{hinton_distilling_2015})}. 

Knowledge distillation traditionally facilitates the transfer of knowledge from a larger, high-performing model to a smaller model, typically of the same architecture, in order to preserve performance while reducing complexity. In this work, we extend the paradigm of knowledge distillation by addressing a novel question: Can knowledge be effectively transferred across architectures, from an attention-based model to a recurrent one?

In this paper, we introduce Distil-xLSTM, a small language model (SLM) based on the xLSTM architecture. Distil-xLSTM leverages the unique capabilities of xLSTM, namely the new memory mixing and parallel computation, to approximate the behavior and performance of an attention-based large language model (LLM). Through cross-architecture distillation, we aim to demonstrate that the expressive power of attention mechanisms can be emulated within a recurrent framework, thus offering a computationally efficient alternative to traditional transformer-based models.

Our approach relies on weight reusing from the teacher and the introduction of a time-varying distillation loss function to help the student model overcome the capacity gap between itself and its teacher. The rest of this paper is structured as follows: we begin with a background section to provide enough context on our work. Then we will provide further details on our methodology and present our experiment's results before continuing with related research in the domain.

\section{Background}

\subsection{Self-attention}
In a transformer-based model, computing attention is a parallel process that uses different parameterized heads (\cite{vaswani_attention_2023}). Given an input $x \in \mathbb{R}^{N \times D}$, a sequence of $N$ vectors of dimensionality $D$, the attention computation is formalized as follows:

\begin{equation}
    \label{eq:self_attention}
    \text{Attention}(Q, K, V) = \text{softmax}\left(\frac{QK^T}{\sqrt{d_k}}\right) V
\end{equation}

where $d_k$ is the dimension of the keys and $Q, K, V$ are respectively the query, key, and value matrices and are obtained by computing $\forall \, x_i \in x, \; Q = W^Qx_i, \; K = W^Kx_i, \; V = W^Vx_i$.

\subsection{The xLSTM architecture}

The Long-Short Term Memory (LSTM) architecture (\cite{hochreiter1997lstm}) is a type of Recurrent Neural Network (RNN) designed to effectively capture long-term dependencies in sequential data by utilizing gating mechanisms that regulate the flow of information, addressing the vanishing gradient problem common in traditional RNNs (\cite{Werbos1990BackpropagationTT}).

The original LSTM formulation is expressed as follows:
\begin{align}
c_t &= f_t + c_{t-1} + i_t z_t && &&\text{cell state} \\
h_t &= o_t \tilde{h}, &&\tilde{h} = \psi(c_t) &&\text{hidden state} \\
z_t &= \varphi(\tilde{z}_t), &&\tilde{z}_t = w_z^\top x_t + r_z h_{t-1} + b_z & &\text{cell input} \\
i_t &= \sigma(\tilde{i}_t), &&\tilde{i}_t = w_i^\top x_t + r_i h_{t-1} + b_i  &&\text{input gate} \\
f_t &= \sigma(\tilde{f_t}) &&\tilde{f}_t = w_f^\top x_t + r_f h_{t-1} + b_f &&\text{forget gate} \\
o_t &= \sigma(\tilde{o_t}) &&\tilde{o}_t = w_o^\top x_t + r_o h_{t-1} + b_o &&\text{output gate}
\end{align}

$w_z, w_i, w_f, w_o$ are input weight vectors between the inputs $x_t$ and the cell input, input gate, forget gate and output gate respectively. $r_z, r_i, r_f, r_o$ are the recurrent weights between the hidden state $h_{t-1}$ and the cell input, input gate, forget gate and output gate respectively. $b_z, b_i, b_f, b_o$ are the corresponding bias terms. $\phi$ and $\psi$ represent the cell input and hidden state activation functions.

Building upon the LSTM architecture, the Extended Long-Short Term Memory (xLSTM) architecture introduces novelties such as a new memory-mixing method through the sLSTM component, as well as a matrix memory and parallel computation using the mLSTM component (\cite{beck_xlstm_2024}). Moreover, the exponential function $\exp{(x)}$ replaces the sigmoid (Equation \ref{eq:sigmoid}) non-linearity, and new states such as normalizer and stabilizer states have been introduced to prevent gradients from overflowing. 

The sLSTM block has a scalar memory and introduces a new memory mixing technique whereas the mLSTM block has a matrix memory and allows parallel computation. The sLSTM forward pass is expressed as:
\begin{align}
    c_t &= f_t c_{t-1} + i_t z_t && &&\text{cell state} \\
    n_t &= f_t n_{t-1} + i_t && &&\text{normalizer state} \\
    h_t &= o_t \tilde{h}, &&\quad \tilde{h} = \frac{c_t}{n_t} &&\text{hidden state} \\
    z_t &= \varphi(\tilde{z}_t), &&\quad \tilde{z}_t = w_z^\top x_t + r_z h_{t-1} + b_z &&\text{cell input} \\
    i_t &= \exp(\tilde{i}_t), &&\quad \tilde{i}_t = w_i^\top x_t + r_i h_{t-1} + b_i &&\text{input gate} \\
    f_t &= \sigma(\tilde{f}_t) \text{ OR } \exp(\tilde{f}_t), &&\quad \tilde{f}_t = w_f^\top x_t + r_f h_{t-1} + b_f &&\text{forget gate} \\
    o_t &= \sigma(\tilde{o}_t), &&\quad \tilde{o}_t = w_o^\top x_t + r_o h_{t-1} + b_o &&\text{output gate}
\end{align}

with
\begin{align}
    m_t &= \max\left(\log(f_t) + m_{t-1}, \log(i_t)\right) & &\text{stabilizer state} \\
    i_t' &= \exp\left(\log(i_t) - m_t\right) = \exp\left(\tilde{i}_t - m_t\right) & &\text{stabilizer input gate} \\
    f_t' &= \exp\left(\log(f_t) + m_{t-1} - m_t\right) & &\text{stabilizer forget gate}
\end{align}

As for the mLSTM block, the LSTM memory represented by a $c$ is increased to a matrix $C \in \mathbb{R}^{d \times d}$ and its forward pass is defined as:

\begin{align}
    \bm{C}_t &= f_t \bm{C}_{t-1} + i_t \bm{v}_t \bm{k}_t^\top && \text{cell state} \\
    \bm{n}_t &= f_t \bm{n}_{t-1} + i_t \bm{k}_t && \text{normalizer state} \\
    \bm{h}_t &= \bm{o}_t \odot \tilde{\bm{h}}_t , \quad \tilde{\bm{h}}_t = \bm{C}_t \bm{q}_t / \max\left\{\left|\bm{n}_t^\top \bm{q}_t\right|, 1\right\} && \text{hidden state} \\
    \bm{q}_t &= W_q x_t + b_q && \text{query input} \\
    \bm{k}_t &= \frac{1}{\sqrt{d}} W_k x_t + b_k && \text{key input} \\
    \bm{v}_t &= W_v x_t + b_v && \text{value input} \\
    i_t &= \exp(\tilde{i}_t), \quad \tilde{i}_t = w_i^\top x_t + b_i && \text{input gate} \\
    f_t &= \sigma(\tilde{f}_t) \, \text{OR} \, \exp(\tilde{f}_t), \quad \tilde{f}_t = w_f^\top x_t + b_f && \text{forget gate} \\
    o_t &= \sigma(\tilde{o}_t), \quad \tilde{o}_t = W_o x_t + b_o && \text{output gate}
\end{align}

\begin{equation}
\sigma (x) = \frac{1}{1 + \exp{(-x)}} \label{eq:sigmoid}
\end{equation}

\subsection{Knowledge distillation}
As originally introduced, knowledge distillation (\cite{hinton_distilling_2015}) aims to transfer knowledge from a bigger and more capable model (generally called teacher model) to a smaller one (the student model) by adjusting the smaller model's logits to match the bigger ones. To do so, during training the loss function is defined as:

\begin{equation}
    \label{eq:distillation_loss}
    \mathcal{L}_{\text{KD}} = (1 - \alpha) \cdot \mathcal{H}(y, z_s) +  \alpha \cdot T^2 \cdot \text{KL}(p_t^{(T)} \vert \vert  p_s^{(T)})
\end{equation}

\begin{equation}
\label{eq:cross_entropy}
    \mathcal{H}(y,z_s) = - \frac{1}{N} \sum_{i=1}^{N} \sum_{j=1}^{C} y_{i,j} \log \left( \frac{\exp(z_{s,i,j})}{\sum_{k=1}^{C} \exp(z_{s,i,k})} \right)
\end{equation}

Where $\mathcal{H}(y, z_s)$ is the cross-entropy loss with $y$ being target labels, $z_s$ the student model's predictions, $N$ the number of samples in the batch, and $C$ the number of classes. $\text{KL}(p_t^{(T)} \vert \vert  p_s^{(T)})$ is the Kullback-Leibler divergence between the softened probability distributions of the teacher model $p_t^{(T)}$ and the student's one $p_s^{(T)}$. $T$ is the temperature used to soften the distributions before computing the Kullback-Leibler divergence and $\alpha$ is a coefficient weighing the importance to give to each term of the loss.

Generally speaking, with knowledge distillation, a smaller version of a capable model is trained from scratch while using the capable one as a reference to align the student model's output to imitate the teacher model output distribution.

\section{Related work}

The Born-Again Multi-task (BAM) framework by \cite{clark2019bambornagainmultitasknetworks} introduced an innovative approach to multitask learning through knowledge distillation. By training a multitask student model using predictions from multiple single-task teacher models, BAM leveraged a mechanism called \textit{teacher annealing}. Early in training, the student heavily relies on the teacher’s guidance. Still, as training progresses, the teacher’s influence is gradually reduced, allowing the student to focus on learning from hard labels. This mechanism effectively balances the benefits of teacher-provided soft targets with independent learning, thereby improving student model performance.

\cite{jafari2021annealingknowledgedistillation} extended this idea with Annealing-KD, focusing on knowledge compression to address the capacity disparity between teacher and student models. Their method dynamically reduced the temperature parameter $T$ after each training epoch, compressing the teacher’s knowledge into a form more suitable for the student’s limited capacity. 

Our proposed $\Delta$-distillation builds upon these ideas but introduces significant innovations. Unlike BAM or Annealing-KD, which focus on either performance improvement or knowledge compression, $\Delta$-distillation simultaneously addresses both. By annealing both the soft target weight ($\alpha$) and temperature ($T$) during and across epochs, our method adapts the teacher-student interaction based on the hypothesis that the student progressively internalizes the teacher's dark knowledge throughout training. This dual annealing mechanism is central to our approach, enabling the student to efficiently assimilate knowledge while achieving improved performance.

A different line of research, exemplified by Bick et al. \cite{bick2024transformersssmsdistillingquadratic}, explores the distillation of transformer-based models into simpler architectures such as Mamba \cite{mamba2}. They introduced the MOHAWK framework, which utilizes a three-stage process combining Matrix Orientation, Hidden state Alignment, Weight-Transfer, and Knowledge Distillation. Their approach included reusing parameters from specific attention blocks of the teacher model. However, their work primarily focused on hybrid models that integrate recurrent and attention-based components.

In contrast, our study is centered on pure recurrent architectures, specifically xLSTMs. While $\Delta$-distillation reuses parameters from the teacher, this is restricted to the embedding layer and classification head, ensuring a strict separation from hybrid design principles. This focus on recurrent models distinguishes our work and highlights the broader applicability of our method for environments where transformers may not be feasible.

Table \ref{tab:comparison_related_work} summarizes the key distinctions between $\Delta$-distillation and related methods, underscoring the novel contributions of our approach.

\begin{table}[h!]
    \centering
    \begin{tabular}{|l|l|l|l|}
        \hline
        \textbf{Method} & \textbf{\makecell{Teacher \\ Architecture}} & \textbf{\makecell{Student \\ Architecture}} & \textbf{Goal} \\
        \hline
        \makecell{Teacher Annealing \\ \cite{clark2019bambornagainmultitasknetworks}} & Transformer & Transformer & 
        \parbox{4cm}{Student performance \\ improvement} \\
        \hline
        \makecell{Annealing-KD \\ \cite{jafari2021annealingknowledgedistillation}} & Transformer & Transformer & 
        \parbox{4cm}{Teacher knowledge \\ compression} \\
        \hline
        \makecell{MOHAWK \\ \cite{bick2024transformersssmsdistillingquadratic}} & Transformer & Mamba/Hybrid & Block-wise matrix alignment \\
        \hline
        $\Delta$-distillation & Transformer & xLSTM & 
        \parbox{4cm}{Student performance improvement \\ and teacher knowledge compression} \\
        \hline
    \end{tabular}
    \caption{Comparison of $\Delta$-distillation with related work}
    \label{tab:comparison_related_work}
\end{table} 

This structured comparison highlights the distinctiveness of $\Delta$-distillation, showcasing its ability to harmonize knowledge compression with robust performance enhancement in purely recurrent architectures.

\section{Key contributions}

In this work, we introduce \textbf{Distil-xLSTM}, an xLSTM-based Small Language Model (SLM) designed to approximate the attention mechanisms of transformer-based models through cross-architecture knowledge distillation. Our main contributions are as follows:

\begin{itemize}
    \item \textbf{Cross-Architecture Distillation}: Demonstration of effective knowledge transfer from a transformer-based teacher to a purely recurrent student architecture (xLSTM). This bridges the gap between attention-based and recurrent models, enabling efficient deployment in resource-constrained settings.

    \item \textbf{Architectural Innovations}: Utilization of xLSTM's enhanced capabilities, including scalar/matrix memory (sLSTM/mLSTM blocks), parallel computation, and stabilizer states to approximate attention mechanisms. The student model employs a reduced yet expressive architecture, initialized with $\sim$50\% fewer layers and optimized head counts derived from the teacher.

    \item \textbf{Frobenius Norm Regularization}: Introduction of a hidden state alignment loss term to compress and stabilize knowledge transfer. This aligns the student's latent representations with the teacher's, mitigating architectural disparities and improving training stability.

    \item \textbf{Computational Efficiency}: Achieved through weight reuse (embedding layer and classification head from the teacher) and minimal trainable parameters (15\% of total), reducing training costs. Experiments on 512M tokens with 551M parameters show convergence comparable to transformer baselines, despite linear recurrent scaling.
\end{itemize}

\section{$\Delta$-Distillation process}

\begin{figure}[t!]
    \centering
    \includegraphics[width=1\linewidth]{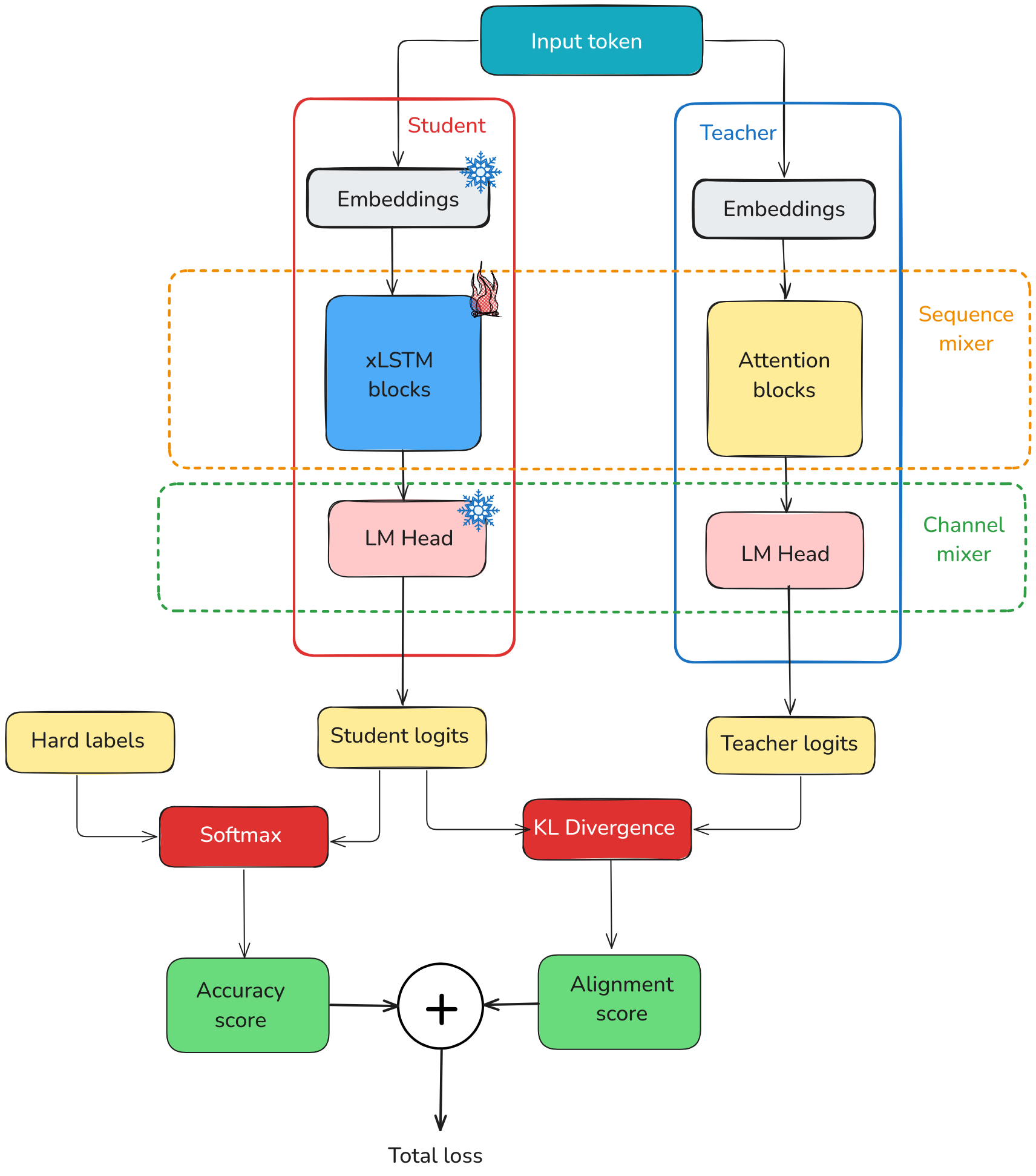}
    \caption{Our distillation framework with frozen embedding layer and classification head initialized using the teacher's weights.}
    \label{fig:distil_xlstm_framework}
\end{figure}

Modern state-of-the-art language models can be deconstructed into three main components: an \textbf{embedding layer}, \textbf{attention blocks} (responsible for sequence mixing), and the \textbf{classification head} (serving as the channel mixer) \cite{bick2024transformersssmsdistillingquadratic}. Central to their success are the attention blocks, where the model learns intricate relationships between tokens, effectively capturing dependencies within the input sequence. Inspired by this framework of sequence and channel mixers, we hypothesize that a recurrent model, specifically one based on xLSTM can approximate the internal representations generated by the attention layers of a transformer.

This hypothesis builds on the work of Katharopoulos et al. \cite{katharopoulos_transformers_2020}, who demonstrated that any transformer layer with causal masking can be reformulated as a recurrent model, with recurrence viewed temporally. Their insight reframes self-attention operations into a row-wise computation, suggesting a path for modeling attention mechanisms through recurrent architectures. By linearizing attention using kernel-based approaches, they laid the foundation for approximating attention mechanisms without explicitly relying on quadratic complexity.

Leveraging this prior, our distillation framework (illustrated in Figure \ref{fig:distil_xlstm_framework}) adopts a novel approach: reusing the teacher model’s embedding layer and classification head weights to initialize their counterparts in the student model. This initialization assumes that the teacher’s parameters for these components are already well-optimized. Consequently, our primary focus shifts to approximating the teacher’s sequence mixer, its attention blocks exclusively through xLSTM blocks. This design simplifies the distillation process while ensuring that the student retains the ability to replicate the teacher's rich internal representations.

By emphasizing the recurrent formulation, our framework not only bridges the gap between transformer-based and recurrent architectures but also demonstrates the feasibility of achieving transformer-level performance with more computationally efficient recurrent models.

To address the challenge of distilling knowledge from a transformer with many attention blocks into an xLSTM model with fewer recurrent layers, we introduce a novel framework called \textbf{$\Delta$-distillation}. This method redefines the traditional knowledge distillation paradigm by employing a time-varying loss function, where the scaling parameters $\alpha$ and $T$ are progressively reduced throughout training. This gradual adjustment encourages the student to rely initially on the teacher's dark knowledge and progressively shift its learning focus to the hard labels provided by the dataset.
The core principles of $\Delta$-Distillation are as follows:

\noindent \textbf{Dual Annealing.} Both $\alpha$ and $T$ are annealed within each epoch using a logarithmic schedule, ensuring a smooth decay that supports stable gradient flows. Across epochs, they are further reduced by a constant factor $\Delta$, making the student less dependent on the teacher over time.

\noindent \textbf{Logarithmic Schedule.} The parameter $\alpha_k$ at a given global training step $k$ is computed using the following schedule:

\begin{equation} \label{eq:annealing_schedule_fn}
\alpha_k = \alpha_{\text{final}} + \frac{\alpha_{\text{initial}} - \alpha_{\text{final}}}{1 + \log{(k + 1)}}
\end{equation}

Similarly, the temperature $T_k$ follows the same schedule, ensuring that the logits remain appropriately softened during the early stages of training and gradually sharpen as training progresses.

\noindent \textbf{Epoch-Wise Decay.} At the end of each epoch, $\alpha$ and $T$ are reduced by a constant factor $\Delta$, ensuring a systematic reduction over the entire training period:

\begin{align}
\alpha &\gets \max(\alpha - \Delta\alpha, \alpha_{\text{final}}) \\
T &\gets \max(T - \Delta T, T_{\text{final}})
\end{align}

\noindent \textbf{Convergence Analysis.} The limit of the schedule function as $k \to +\infty$ is given as follows:

%\begin{align} \label{eq:schedule_fn_limit}
%\lim_{k \to +\infty} \alpha_k &= \alpha_{\text{final}}
%\end{align}

 \begin{align} \label{eq:schedule_fn_limit}
 \lim_{k \to +\infty} \alpha_k &= \lim_{k \to +\infty} \left( \alpha_{\text{final}} + \frac{\alpha_{\text{initial}} - \alpha_{\text{final}}}{1 + \log{(k + 1)}} \right) \\
 &=\alpha_{\text{final}} + \underbrace{\lim_{k \to +\infty} \frac{\alpha_{\text{initial}} - \alpha_{\text{final}}}{1 + \log{(k + 1)}}}_{=0} \\
 &= \alpha_{\text{final}}
 \end{align}

This ensures that $\alpha_k$ and $T_k$ stabilize at their respective final values, enabling the student to continue receiving minimal guidance while learning from hard labels.

\noindent \textbf{Time-Varying Loss Function.} Another central component of $\Delta$-distillation is its time-varying loss function that progressively changes within and across each epoch. The student model's loss is a weighted sum of two components: (1) the knowledge distillation loss $\mathcal{L}_{\text{KD}}$, which uses softened logits from the teacher, and (2) the cross-entropy loss $\mathcal{L}_{\text{CE}}$ (same as Equation \ref{eq:cross_entropy}), based on the hard labels from the dataset. 

This combined loss is defined as:

\begin{equation} \label{eq:varying_distillation_loss}
\mathcal{L}_{\text{distill}} = (1 - \alpha_k) \cdot \mathcal{L}_{\text{CE}} + \alpha_k \cdot T_{k}^2 \cdot \mathcal{L}_{\text{KD}}(T_k)
\end{equation}

Here:
\begin{itemize}
    \item $\alpha_k \in [0, 1]$: Determines the weight given to the teacher’s guidance.
    \item $T_k > 0$: The temperature scalar used to soften the logits from the teacher.
\end{itemize}

%\subsection{Algorithm Description}

The distillation process is formalized in Algorithm~\ref{algo:delta_distillation}.

To further mitigate the capacity gap between the student and the teacher, we dynamically initialize the student model's sequence mixer using the following heuristic:

\begin{enumerate}
    \item \textbf{Number of Sequence Mixing Layers}:
    Let $L_t$ be the number of attention layers in the teacher's sequence mixer. The student's sequence mixer is initialized with 
    $
    L_s = \left\lfloor \frac{L_t}{2} \right\rfloor
    $
    xLSTM layers, where $\left\lfloor \cdot \right\rfloor$ denotes the floor function. This ensures that the student has approximately half the depth of the teacher, reducing the capacity gap while maintaining computational efficiency.

    \item \textbf{Number of Heads}:
    Let $H_t$ be the number of attention heads within each attention layer of the teacher. Each xLSTM layer in the student model is initialized with 
    $
    H_s = \text{roundup}(H_t, 4),
    $
    where $\text{roundup}(x, k)$ rounds $x$ up to the nearest multiple of $k$. This ensures that the number of heads in the student's xLSTM layers is both expressive and computationally efficient.
\end{enumerate}

% \subsection*{Example}

% \begin{itemize}
%     \item \textbf{Teacher Model}:
%     \begin{itemize}
%         \item $L_t = 24$ (24 attention layers).
%         \item $H_t = 16$ (16 attention heads per layer).
%     \end{itemize}

%     \item \textbf{Student Model}:
%     \begin{itemize}
%         \item $L_s = \left\lfloor \frac{24}{2} \right\rfloor = 12$ (12 xLSTM layers).
%         \item $H_s = \text{roundup}(16, 4) = 16$ (16 heads per xLSTM layer).
%     \end{itemize}
% \end{itemize}

By doing so, we address the following points:
\begin{itemize}
    \item \textbf{Capacity Matching}: By setting $L_s = \left\lfloor \frac{L_t}{2} \right\rfloor$, the student model has approximately half the depth of the teacher, which helps bridge the capacity gap without making the student too large or computationally expensive.

    \item \textbf{Expressive Attention Mechanisms}: By setting $H_s = \text{roundup}(H_t, 4)$, the student’s xLSTM layers have a sufficient number of heads to mimic the teacher’s attention mechanisms while ensuring computational efficiency.

    \item \textbf{Dynamic Initialization}: The heuristic dynamically adjusts the student’s architecture based on the teacher’s configuration, making it adaptable to different teacher models.
\end{itemize}

\begin{algorithm}[H]
\caption{$\Delta$-Distillation Framework}
\label{algo:delta_distillation}
    \begin{algorithmic}[1]
        \State \textbf{Input:} $\alpha_{\text{initial}}, T_{\text{initial}}, \, \alpha_{\text{final}}, \, T_{\text{final}}, \, \Delta \alpha, \, \Delta T, \, n_{\text{epochs}}, \, \text{steps\_per\_epoch}$
        \For{epoch $= 1$ to $n_{\text{epochs}}$}
            \For{step $= 1$ to $\text{steps\_per\_epoch}$}
                \State Perform forward pass and compute the distillation loss:
                \[ \mathcal{L}_{\text{distill}} = (1 - \alpha_k) \cdot \mathcal{L}_{\text{CE}} + \alpha_k \cdot \mathcal{L}_{\text{KD}}(T_k) \]

                \State Perform backward pass and update model parameters
                \State Update $\alpha_k$ using the schedule:
                \[ \alpha_k \gets \alpha_{\text{final}} + \frac{\alpha - \alpha_{\text{final}}}{1 + \log{(\text{step} + 1)}} \]

                \State Update $T_k$ using the same schedule
            \EndFor
            \State Update $\alpha$ for the next epoch:
            $\alpha \gets \max(\alpha - \Delta\alpha, \alpha_{\text{final}})$
            \State Update $T$ for the next epoch:
            $T \gets \max(T - \Delta T, T_{\text{final}})$
        \EndFor
    \end{algorithmic}
\end{algorithm}

%\subsection{Key Benefits}
The key benefits of our approach can be summarised as follows:

% \begin{itemize}
%     \item \textbf{Progressive knowledge transfer:} By dynamically annealing $\alpha$ and $T$, the framework ensures the student transitions smoothly from relying on the teacher’s soft labels to hard labels, reducing overfitting risks.
%     \item \textbf{Efficient knowledge compression:} The use of xLSTM blocks to approximate transformer attention mechanisms minimizes the parameter footprint while retaining robust performance.
%     \item \textbf{Stable training dynamics:} The logarithmic decay offers a controlled reduction, avoiding abrupt changes that can destabilize training.
% \end{itemize}

\begin{itemize}
    \item Dynamic teacher-student balance: By gradually transitioning from teacher-guided knowledge distillation to self-reliant learning, the combined loss ensures the student model effectively captures both the teacher's expertise and the dataset's inherent structure.

\item Improved generalization: The balance between $\mathcal{L}_{\text{KD}}$ (softened logits) and $\mathcal{L}_{\text{CE}}$ (hard labels) prevents overfitting to either the teacher's dark knowledge or the dataset, leading to better generalization on unseen data.

\item  Smooth learning transition: The progressive adjustment of $\alpha$ and $T$ enables a stable and controlled shift in the learning focus, avoiding abrupt changes that could destabilize the optimization process.

\item  Knowledge compression with confidence calibration: By using a softened output (via $T$) during early training, the framework enables the student to learn rich, nuanced representations, while gradually sharpening logits ensures confident and decisive predictions as training progresses.
\end{itemize}

The $\Delta$-distillation framework achieves an effective balance between teacher guidance and independent learning, enabling efficient distillation into compact and recurrent architectures.

\section{Experimental results}
We trained the proposed Distil-xLSTM model using Qwen2.5-1.5B (\cite{qwen2024qwen25technicalreport}) as the teacher model. For efficiency, experiments were conducted on an Nvidia A100 GPU with FP16 mixed precision training (\cite{micikevicius2018mixedprecisiontraining}). The training was performed on $512M$ tokens from the FineWeb dataset (\cite{penedo2024finewebdatasetsdecantingweb}) over 10 epochs. 

By reusing the embedding layer and classification head weights from the teacher model, our Distil-xLSTM architecture consists of six xLSTM blocks, alternating between sLSTM and mLSTM blocks in a $1:1$ ratio, starting with an sLSTM block. The model contains 551M parameters, of which only $84$M ($\approx 15.24\%$), corresponding to the sequence mixer's parameters, are trainable. This significantly reduces the training cost while preserving performance. A summary of the training hyperparameters is presented in Table \ref{tab:training_hparams}.

\begin{table}[H]
    \centering
    \begin{tabular}{|l|l|}
        \hline
        Hyperparameter & Value\\
        \hline
        Learning rate & $2 \cdot10^{-4}$\\
        \hline
        Learning rate scheduler & Cosine\\
        \hline
        Batch size & 8\\
        \hline
        Gradient accumulation & 4\\
        \hline
        Warmup ratio & 0.1\\
        \hline
        $\alpha_{\text{initial}}$ & 0.8\\
        \hline
        $\alpha_{\text{final}}$ & 0.5\\
        \hline
        $T_{\text{initial}}$ & 2\\
        \hline
        $T_{\text{final}}$ & 1\\
        \hline
        $\Delta$ & 0.05 \\
        \hline
        Context size & 512 tokens \\
        \hline
    \end{tabular}
    \caption{Hyperparameters used to train Distil-xLSTM}
    \label{tab:training_hparams}
\end{table}

\subsection{Regularization with the Frobenius Norm}
The main objective of this research is to align an xLSTM-based model's sequence mixer parametrization with an attention-based one. In addition to our proposed $\Delta$-distillation, we empirically find that adding the Frobenius norm to the loss encourages the student's hidden state to align with the teacher's hidden state in terms of magnitude and representation, with nearly no loss in performance compared to $\Delta$-distillation. This regularization provides the following benefits:

\begin{enumerate}
    \item \textbf{Aligning Hidden States:} The Frobenius norm encourages the student's hidden state to align with the teacher's hidden state in terms of magnitude and representation. It minimizes the difference between the two latent representations, ensuring that the student network approximates the teacher's internal feature extraction process.
      
\item \textbf{Compressing Representations:} Since the student model (xLSTM) has fewer sequence mixing blocks and a different architecture compared to the teacher model (transformer), the Frobenius norm provides a mechanism for knowledge transfer across these disparate architectures. By focusing on aligning the latent representations, the student learns to emulate the teacher's knowledge efficiently, despite having fewer parameters.
      
\item \textbf{Regularizing Distillation:} The Frobenius norm acts as a regularizer for the knowledge distillation process. It ensures that the student network does not deviate too far from the teacher's hidden representations, helping to stabilize training. This is particularly useful when combining different loss terms like cross-entropy (CE) and Kullback-Leibler (KL) divergence.
\end{enumerate}

Let $x \in \mathbb{R}^{B \times S}$ be a batch of $B$ input sequences of size $S$. The xLSTM produces a final single state $h_s \in \mathbb{R}^{B \times S \times D}$ before outputting logits. To benefit the most from the transformer's expressivity, we retrieve the hidden state produced by each attention layer and stack them to form a global hidden state $h_t \in \mathbb{R}^{B' \times S \times D}$, where $B' = L_t \times B$. By transforming $h_t$ to match the shape $L_t \times B \times S \times D$, we can compute a layer-wise average of hidden states $\bar{h}_t \in \mathbb{R}^{B \times S \times D}$ that matches the student model's hidden state shape.

Our new loss term is defined as follows:
    \begin{equation}
        \label{eq:frobenius_norm_as_loss}
        \mathcal{L}_{\text{frobenius}} = \frac{1}{B} \sum_{i=1}^{B} \| \bar{h}_{\text{t}}^{(i)} - h_{\text{s}}^{(i)} \|_F,
    \end{equation}
    
where $B$ is the batch size, and $h_t^{(i)}$ is the hidden state produced for the $i$-th input sequence. To stabilize the contribution of the Frobenius norm to the distillation loss function, we normalize it by dividing by $\sqrt{|h_s|}$, where $|h_s|$ denotes the number of elements in the tensor $h_s$.

By introducing an additional scalar $\beta$ to weight the contribution of the Frobenius norm, the distillation loss function (Eq. \ref{eq:varying_distillation_loss}) can be rewritten as:
    \begin{equation}
        \label{eq:distillation_loss_with_frobenius}
        \mathcal{L}_{\text{distill}} = (1 - \alpha - \beta) \cdot \mathcal{L}_{\text{CE}} + \alpha \cdot T^2 \cdot \mathcal{L}_{\text{KD}}(T) + \beta \cdot \frac{\mathcal{L}_{\text{frobenius}}}{\sqrt{|h_s|}}.
    \end{equation}

With respect to the idea of $\Delta$-distillation, we proceeded to further experiments by applying its annealing scheme to Eq. \ref{eq:distillation_loss_with_frobenius} and obtained results almost similar to the distillation process with fixed scalars. To avoid having the overall loss dominated by the Frobenius norm, we weigh it with $\beta = 0.1$ and set $\alpha = 0.3$. For the time-varying form, we have $\beta_k \in [0.1, 0.2]$ and $\alpha \in [0.2, 0.3]$.

\subsection{Training Results}

\begin{figure}[H]
    \centering
    \includegraphics[width=1\linewidth]{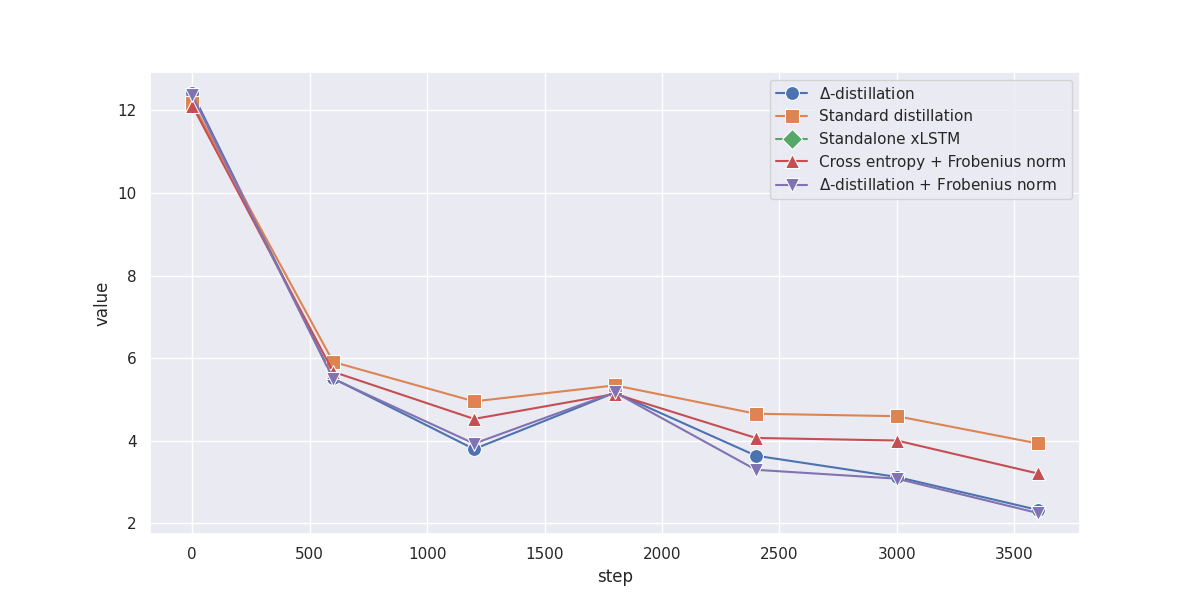}
    \caption{$\mathcal{L}_{\text{CE}}$ during training}
    \label{fig:train_ce_loss}
\end{figure}

\begin{figure}[H]
    \centering
    \includegraphics[width=1\linewidth]{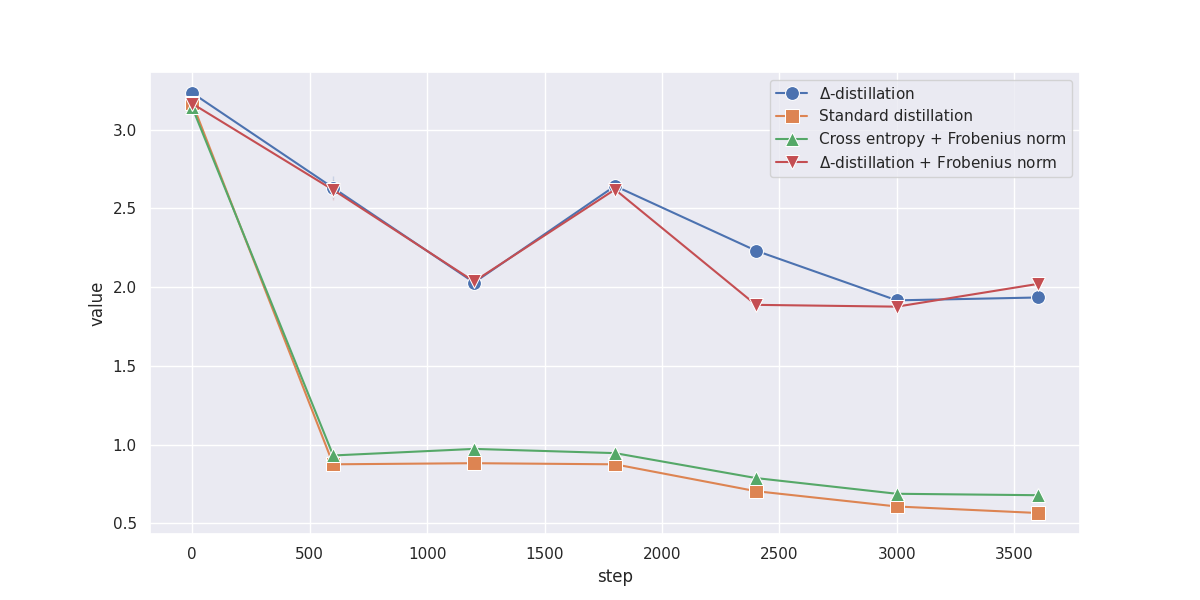}
    \caption{$\mathcal{L}_{\text{KL}}$ during training}
    \label{fig:train_kl_loss}
\end{figure}

\begin{figure}[H]
    \centering
    \includegraphics[width=1\linewidth]{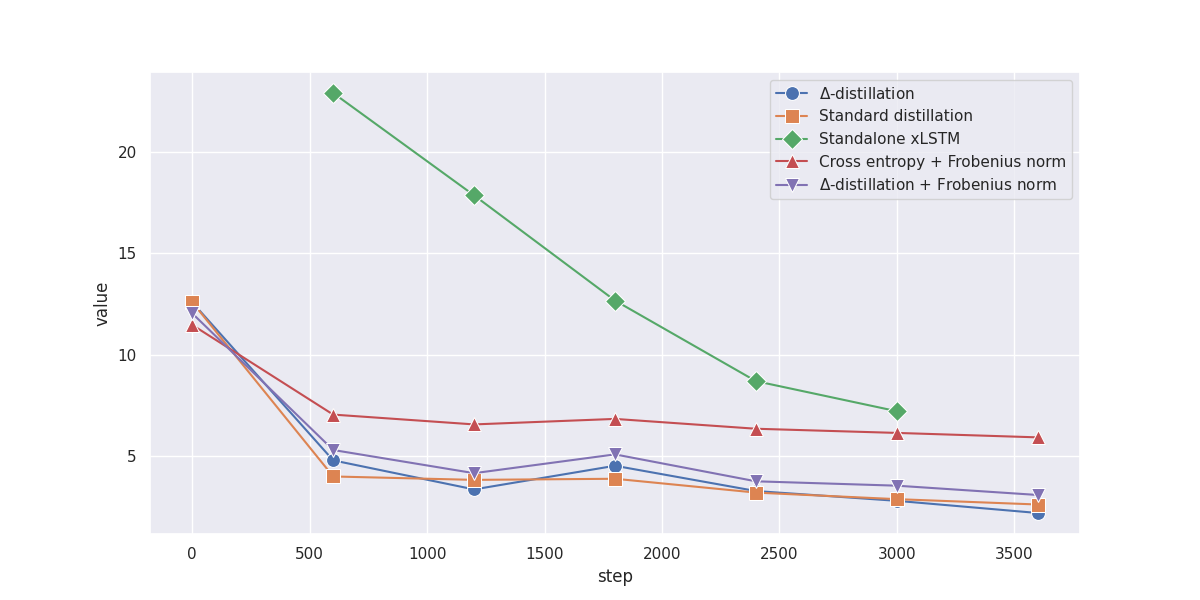}
    \caption{Overall loss ($\mathcal{L}_{\text{distill}}$) during training}
    \label{fig:train_total_loss}
\end{figure}

Figure \ref{fig:train_total_loss} shows the convergence of the training loss over 10 epochs, indicating effective learning. The cross-entropy loss (Figure \ref{fig:train_ce_loss}) also decreases steadily, demonstrating the model's ability to learn from hard labels. Oscillations related to the Kullback-Leibler divergence (Figure \ref{fig:train_kl_loss}) over time are due to the fact that the model progressively shifts its emphasis from the knowledge provided by the teacher to focus on training data. In the beginning, this loss's landscape is decreasing, confirming that the student is successfully distilling knowledge from the teacher model.

The decrease in both cross-entropy and KL divergence indicates that the student model effectively balances learning from ground truth labels and teacher guidance. The low percentage of trainable parameters highlights the efficiency of our approach.

\begin{figure}[H]
    \centering
    \includegraphics[width=1\linewidth]{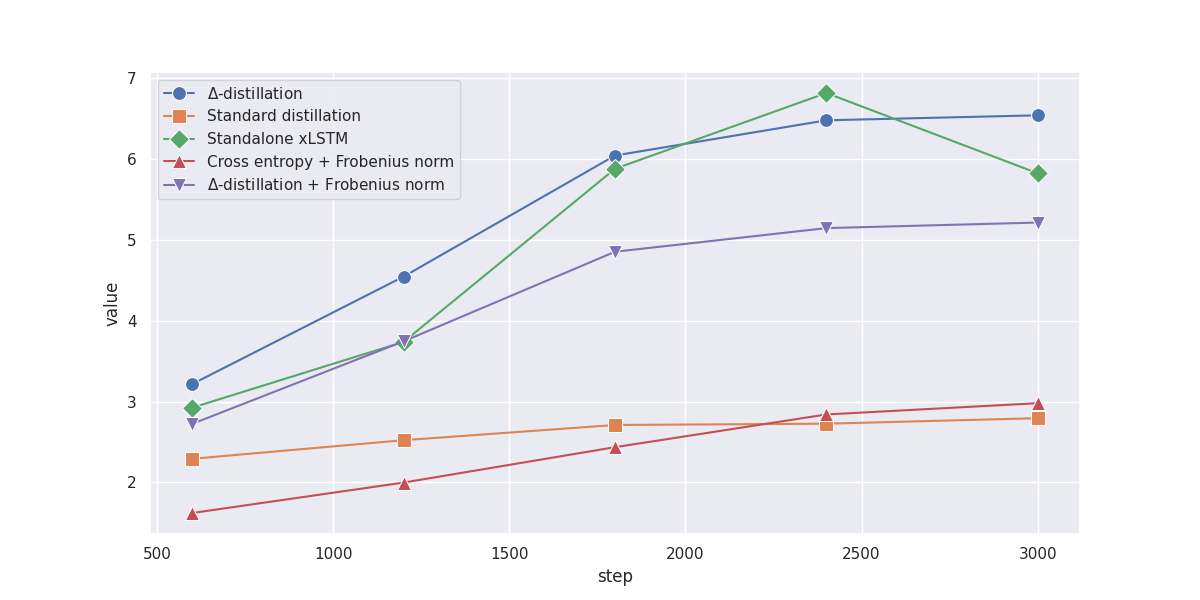}
    \caption{Gradients norm during training}
    \label{fig:train_grad_norm}
\end{figure}

In addition, the addition of the Frobenius norm further stabilized the training process. The model trained with this regularization term not only shows performance similar to the one trained with our proposed $\Delta$-distillation, and it does so while requiring less important updates as it can be seen with the gradients' norm (Fig. \ref{fig:train_grad_norm}). Hence, this empirically validates that the Frobenius norm is an effective component to our distillation framework.

\begin{figure}[H]
    \centering
    \includegraphics[width=1\linewidth]{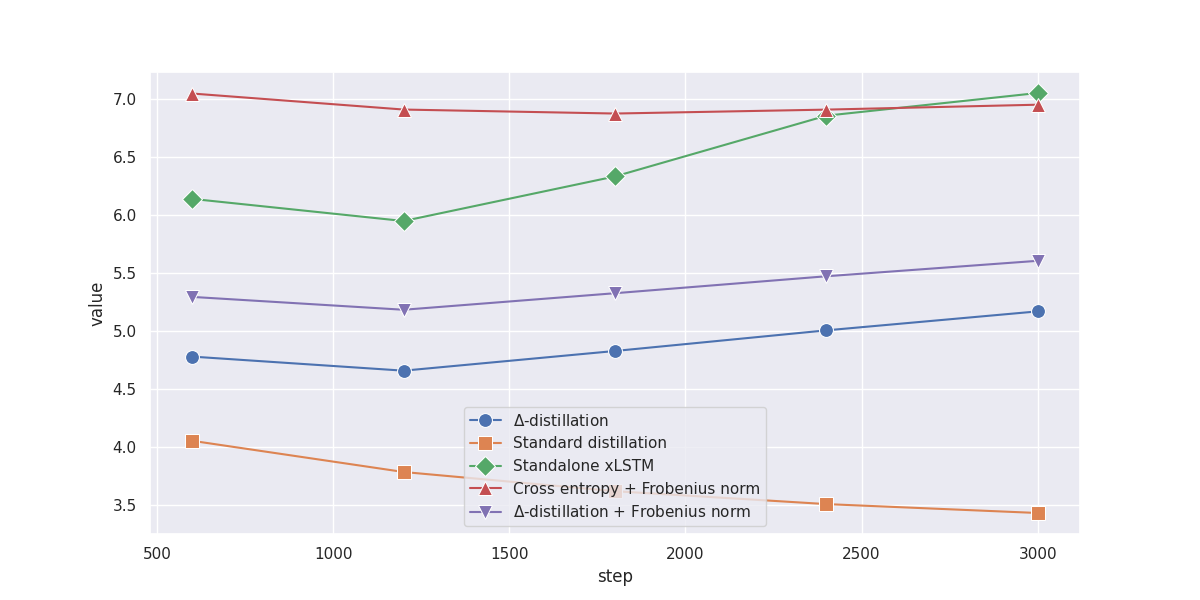}
    \caption{Evaluation loss}
    \label{fig:eval_loss}
\end{figure}

\section{Conclusion}

In this work, we presented Distil-xLSTM, a novel approach for learning attention mechanisms within a recurrent framework through knowledge distillation from a transformer-based teacher model. By leveraging the xLSTM architecture and introducing $\Delta$-distillation, we demonstrated that it is possible to approximate the expressive power of attention mechanisms while maintaining computational efficiency. Our experiments, conducted on a small-scale dataset with constrained computational resources, validate the potential of our method to bridge the gap between attention-based and recurrent models.

While the results are promising, they were achieved on a limited scale due to resource constraints. As part of our future work, we aim to scale up our experiments to larger datasets and more complex tasks, which will further test the robustness and generalizability of Distil-xLSTM. We believe this direction holds significant promise for environments requiring efficient yet capable models, particularly in resource-constrained settings.

%\clearpage 
%\bibliographystyle{abbrv}
\bibliographystyle{spbasic_updated}
\bibliography{main}
\end{document}